\documentclass[pdflatex,sn-mathphys-num]{sn-jnl}% Math and Physical Sciences Numbered Reference Style
\usepackage{graphicx}%
\usepackage{multirow}%
\usepackage{amsmath,amssymb,amsfonts}%
\usepackage{amsthm}%
\usepackage{mathrsfs}%
\usepackage[title]{appendix}%
\usepackage{xcolor}%
\usepackage{textcomp}%
\usepackage{manyfoot}%
\usepackage{booktabs}%
\usepackage{algorithm}%
\usepackage{algorithmicx}%
\usepackage{algpseudocode}%
\usepackage{listings}%
\usepackage{bbm}
\usepackage{placeins}
\theoremstyle{thmstyleone}%
\theoremstyle{thmstyletwo}%

\theoremstyle{thmstylethree}%

\makeatletter
\newenvironment{breakablealgorithm}
  {%
   \begin{center}
   \refstepcounter{algorithm}%
   \hrule height .8pt depth 0pt \kern 2pt
   \renewcommand{\caption}[2][\relax]{%
     {\raggedright\textbf{Algorithm \thealgorithm} ##2\par}%
     \ifx\relax##1\relax
       \addcontentsline{loa}{algorithm}
       {\protect\numberline{\thealgorithm}##2}%
     \else
       \addcontentsline{loa}{algorithm}
       {\protect\numberline{\thealgorithm}##1}%
     \fi
     \kern 2pt
     \hrule
     \kern 2pt
   }%
  }
  {%
   \kern 2pt
   \hrule
   \relax
   \end{center}
  }
\makeatother

\begin{document}

\title[FIBER: Bayesian Fusion of Fast-DetectGPT and IRT-Informed Detector Evidence for LLM-Generated Text Detection]{FIBER: Bayesian Fusion of Fast-DetectGPT and IRT-Informed Detector Evidence for LLM-Generated Text Detection}


\author[1]{\fnm{Yan} \sur{Li}}\email{yanli@sspu.edu.cn}

\author[2]{\fnm{Guangsheng} \sur{Bao}}\email{baoguangsheng@fyust.edu.cn}

\author[3]{\fnm{Zhipeng} \sur{Chen}}\email{chenzhipeng0104@gmail.com}

\author[1]{\fnm{Guandong} \sur{Zhang}}\email{gdzhang@sspu.edu.cn}

\author*[1]{\fnm{Yanbin} \sur{Zhao}}\email{zhaoyb553@nenu.edu.cn}

\affil[1]{\orgdiv{School of Mathematics and Statistics}, \orgname{Shanghai Polytechnic University}, \orgaddress{\street{2360 Jinhai Road}, \city{Pudong New District}, \postcode{201209}, \state{Shanghai}, \country{China}}}

\affil[2]{\orgdiv{School of Computing and Artificial Intelligence}, \orgname{Fuyao University of Science and Technology}, \orgaddress{\street{No. 104 Zhihui Avenue}, \city{Fuzhou High-tech Industrial Development Zone}, \postcode{350109}, \state{Fujian}, \country{China}}}

\affil[3]{\orgdiv{School of Computer and Information Engineering}, \orgname{Shanghai Polytechnic University}, \orgaddress{\street{2360 Jinhai Road}, \city{Pudong New District}, \postcode{201209}, \state{Shanghai}, \country{China}}}

\abstract{Detecting large language model (LLM)-generated text remains challenging under domain and generator shifts. Supervised detectors may generalize poorly, while zero-shot methods often rely on individual statistical signals without leveraging historical detector behavior. We propose FIBER (Fast-DetectGPT with IRT-informed Bayesian Evidence Recalibration), which integrates continuous Fast-DetectGPT scores with auxiliary detector outputs through Bayesian evidence fusion. Item response theory (IRT) models historical detector-correctness patterns, providing class-conditional detector-output likelihoods through marginalization over latent text representations. These likelihoods are combined with Fast-DetectGPT score densities and a historical class prior for prediction. When historical Fast-DetectGPT scores are unavailable, a variant called FIBER-Stream estimates class-specific score distributions from an unlabeled batch. On a 32,000-text multi-domain benchmark, FIBER achieves the highest accuracy (0.620) and improves mean AUC over Fast-DetectGPT (0.664 versus 0.600) under five-fold cross-validation. FIBER also consistently improves AUC across four individual domains. Under out-of-distribution transfer, FIBER-Stream achieved an AUC of 0.911, compared with 0.889 for FIBER and 0.922 for Fast-DetectGPT alone. These results demonstrate the potential of integrating complementary detector evidence and historical information to improve detection performance across heterogeneous settings.}




\maketitle

\section{Introduction}\label{sec1}

Large language models (LLMs) have become widely used in text generation, code synthesis, and multimodal reasoning \citep{brown2020,jiang2024,achiam2023gpt}. As their outputs become increasingly fluent and human-like \citep{chakraborty2023, wu2025}, concerns over content authenticity, misinformation, academic integrity, and responsible AI use have intensified \citep{wu2025,wang2026,ma2025,perkins2023,cotton2024, alharthi2025,porsdam2023}. Detecting LLM-generated text has therefore become increasingly important. However, rapid advances in generative models, together with substantial variation in domains, prompts, decoding strategies, and writing styles,  can induce considerable distribution shifts and limit the generalizability of existing detectors \citep{koike2024, dubois2025}. 

Existing approaches to LLM-generated text detection can be broadly divided into watermarking-based and post-hoc detection methods. Watermarking embeds detectable signals during text generation \citep{gu2022}, but typically requires access to or modification of the generation process, limiting its applicability to arbitrary black-box outputs. We therefore focus on post-hoc detection methods.

Post-hoc supervised detectors formulate the task as binary classification using labeled human-written and machine-generated texts. RoBERTa-based detectors and related pretrained language-model classifiers can achieve strong in-distribution performance \citep{devlin2019,solaiman2019,liu2019}. More recently, AdaDetectGPT \citep{zhou2025} improves the adaptability of logits-based detection by learning a witness function from external labeled data. Despite their effectiveness, supervised approaches generally require labeled examples for model fitting or adaptation and may degrade when the domain, writing style, text length, prompting strategy, or source generator differs from the training distribution \citep{pu2023,wu2025}. 

Zero-shot methods avoid task-specific retraining by exploiting statistical characteristics derived from pretrained language models. Representative approaches use token-level likelihood, entropy, or rank information to identify distributional differences between human-written and machine-generated text \citep{hashimoto2019,gehrmann2019,ippolito2020,su2023}. GLTR \citep{gehrmann2019}, for example, examines token-rank patterns under a reference language model, while DetectLLM \citep{su2023} develops rank-based statistics such as Log-Likelihood Log-Rank Ratio (LRR) and Normalized Perturbed Log-Rank (NPR). Perturbation- and curvature-based methods provide another important family of zero-shot detectors. DetectGPT \citep{mitchell2023} measures changes in log probability after meaning-preserving perturbations, based on the observation that machine-generated texts tend to lie near local maxima of a language model's probability surface. Fast-DetectGPT \citep{bao2023fast} substantially reduces this computational burden by replacing repeated text perturbation with a conditional probability-curvature statistic. Binoculars \citep{hans2024} instead compares probability behavior across two pretrained language models and detects machine-generated text through the discrepancy between their scoring patterns. Although these methods provide effective text-level signals, each relies on a particular statistical perspective, and their relative performance can vary substantially across domains and generators.

Recent studies have therefore explored combining multiple detectors or detection signals. EnsemJudge \citep{wang2026en} integrates rule-based, training-free, and training-based detectors and selects detector subsets and weights according to input characteristics such as text length and perplexity. DWT-Fusion \citep{ozdacs2026} constructs multiple wavelet-based detection scores and combines them through equal-weight or calibration-weighted hard and soft voting. These approaches demonstrate the benefit of exploiting complementary detector information. However, conventional aggregation still has limitations. Equal weighting assumes comparable detector reliability, whereas globally estimated weights reflect average training-set behavior and may become unreliable when relative detector performance changes under distribution shift. More importantly, such approaches make limited use of historical information about how individual texts interact with a collection of detectors.

Historical detector responses contain richer information than detector-level accuracy or global ensemble weights alone. Some texts may be consistently classified correctly by most detectors, whereas others may induce substantial disagreement even within the same class. Such patterns reflect systematic sample-level heterogeneity in detector–text interactions and may provide useful information about the difficulty and discrimination characteristics of individual texts. A natural question is therefore how to summarize these historical interactions and incorporate them into subsequent detection.

To characterize this structure, we adopt item response theory (IRT) \citep{lord1968,lord1980,embretson2013}. IRT was originally developed for educational and psychological measurement to model interactions between respondents with different latent abilities and items with different characteristics. This formulation naturally extends to LLM-generated text detection by treating detectors as respondents, texts as items, and detector correctness as binary responses. The resulting model jointly characterizes detector ability and text-level difficulty and discrimination, providing structured and interpretable summaries of historical detector-text interactions. Importantly, these IRT parameters describe how a text is handled by a collection of detectors rather than directly representing its linguistic or semantic content  \citep{zhao2026}. Further details are given in Section \ref{sub_irt}.

Motivated by the class-conditional patterns observed in these historical IRT representations, we propose Fast-DetectGPT with IRT-informed Bayesian Evidence Recalibration (FIBER). FIBER retains the continuous Fast-DetectGPT score as the primary text-level detection signal and combines it with the current multi-detector output profile and historical detector-response information. Specifically, the class-conditional distributions of historical IRT item parameters are used to construct an IRT-informed detector-output likelihood by integrating over the unobserved item representations. This likelihood is combined with the historical class prior, estimated from the labeled class proportions, and the Fast-DetectGPT score likelihood through Bayesian updating. The resulting posterior probability integrates current text-level evidence with information learned from historical detector behavior. Importantly, test-time inference does not require estimating an item-specific IRT parameter for a new text because the unobserved IRT representation is integrated out using its historical class-conditional distribution.

When labeled historical data are available, the historical class prior, class-conditional IRT distributions, and class-conditional Fast-DetectGPT score distributions can be estimated directly from them. Under compatible detector configurations, the resulting summaries can also serve as reusable historical information for subsequent detection tasks. We further consider a more restrictive setting in which the original historical texts and their Fast-DetectGPT scores are unavailable. To address this setting, we develop FIBER-Stream, which retains historical labels, detector outputs, and IRT summaries while estimating the unavailable class-conditional Fast-DetectGPT score distributions from incoming unlabeled batches through an expectation-maximization procedure. This design allows structured previously learned detector-response information to remain useful even when the original historical texts cannot be revisited.

Our contributions are summarized as follows. 
\begin{itemize}

\item \textbf{A Bayesian evidence-recalibration framework built on Fast-DetectGPT.}
We propose \textbf{FIBER}, which integrates the continuous Fast-DetectGPT score with complementary evidence from multiple detector outputs and historical detector-response information through Bayesian updating. Rather than relying on a single detector or conventional detector-level voting, FIBER combines heterogeneous evidence to estimate the posterior probability that a text is machine-generated.

\item \textbf{An IRT-based summary of historical detector-text interactions.}
We apply multidimensional item response theory to historical detector correctness patterns to characterize text-level difficulty and discrimination. The fitted item representations are subsequently summarized through class-conditional distributions and incorporated into an IRT-informed detector-output likelihood by marginalizing over the unobserved item representations. This provides a structured way to include historical detector-response information in Bayesian evidence recalibration without requiring test-specific IRT estimation.

\item \textbf{Reusable historical information for cross-setting detection.}
We construct a balanced benchmark containing 32,000 human-written and machine-generated texts across multiple domains and generators and evaluate a diverse set of supervised and zero-shot detectors. Beyond detector predictions, the benchmark provides correctness-response matrices, fitted IRT item parameters, detector-output likelihoods, and class-prior estimates that can serve as reusable historical information under compatible detection settings.

\item \textbf{A summary-data extension through FIBER-Stream.}
We develop \textbf{FIBER-Stream} for settings in which the original historical texts and their Fast-DetectGPT scores are unavailable. By combining retained historical summaries with EM-based estimation from incoming unlabeled batches, FIBER-Stream reduces historical-data requirements while preserving the Bayesian evidence-integration structure of FIBER.
\end{itemize}

The remainder of this article is organized as follows. Section \ref{sec2} introduces the detection task and relevant preliminary methods. Section \ref{sec3} presents FIBER and FIBER-Stream, with detailed algorithms provided in the Appendix. Section \ref{sec4} reports the experimental results, and Section \ref{sec5} concludes the article.

\section{Preliminaries}\label{sec2}
\subsection{Task setting}
The goal of LLM-generated text detection is to determine whether a given text is human-written or machine-generated. We formulate this task as a binary classification problem. For a text $X$, let $Y\in \{0,1\}$ denote its true label, where $Y=0$ represents human-written text and $Y=1$ represents machine-generated text. The objective is to estimate the probability that $X$ is machine-generated and make a final prediction according to a prespecified decision threshold.

We consider a post-hoc detection setting in which the source generator of a test text is unknown and need not be accessible. Detection is based on information obtained from the observed text and a collection of existing detectors. In addition to incoming texts, labeled historical samples may be available and can provide information about past detector behavior. The proposed framework combines such historical information with current detection evidence; the corresponding data structures and inference procedures are introduced in Section \ref{sec3}.

\subsection{DetectGPT and Fast-DetectGPT}\label{sub_fast}

\textbf{DetectGPT.} DetectGPT \citep{mitchell2023} is a zero-shot detection method based on the probability curvature of a text under a reference language model. Its key observation is that machine-generated texts tend to lie near local maxima of the model's log probability surface. Consequently, small meaning-preserving perturbations of a machine-generated text often receive lower likelihoods than the original text. 

Given an input text $X$, DetectGPT generates a collection of perturbed texts $\{\widetilde{X}^{(m)}\}_{m=1}^M$ using a pretrained mask-filling model and compares their log probabilities with that of the original text under a scoring model $p_\theta$. A normalized discrepancy score can be written as
$$\Delta(X)=\frac{\log p_\theta(X)-\frac{1}{M}\sum_{m=1}^M\log p_\theta(\widetilde{X}^{(m)})}{\sqrt{\frac{1}{M-1}\sum_{m=1}^M(\log p_\theta(\widetilde{X}^{(m)})-\frac{1}{M}\sum_{l=1}^M\log p_\theta(\widetilde{X}^{(l)}))^2}}.$$

A larger discrepancy provides stronger evidence that the input is machine-generated. DetectGPT does not require task-specific labeled training data, but generating and rescoring multiple perturbations for each text can be computationally expensive.

\textbf{Fast-DetectGPT.} Fast-DetectGPT \citep{bao2023fast} replaces explicit text perturbation with token-level conditional probability calculations, substantially reducing the computational cost of curvature-based detection. Let $p_\theta$ denote the scoring model and $q_\psi$ the sampling model. For a text $X=(x_1,\cdots, x_T)$, Fast-DetectGPT evaluates how the log probability of each observed token differs from the expected log probability of alternative tokens drawn from the sampling distribution.

A simplified form of the normalized conditional probability-curvature score can be expressed as
$$s(X)=\frac{\sum_{t=1}^T\left[\log p_\theta(x_t|x_{<t})-\mathbb{E}_{\tilde{x}_t\sim q_{\psi}(\cdot|x_{<t})}\log p_\theta(\widetilde{x}_t|x_{<t})\right]}{\sqrt{\sum_{t=1}^T\left(Var_{\tilde{x}_t\sim q_{\psi}(\cdot|x_{<t})}[\log p_\theta(\widetilde{x}_t|x_{<t})]\right)}}.$$

The score measures how strongly the observed token sequence deviates from the conditional alternatives implied by the sampling model, after normalization by their variability. Larger scores generally provide stronger evidence of machine generation. In FIBER, this continuous Fast-DetectGPT score is retained directly as the primary text-level detection signal rather than being converted into a binary decision.

\subsection{Item response theory}\label{sub_irt}
Item response theory (IRT) provides a statistical framework for modeling binary response data. In conventional applications, the probability of a correct response depends jointly on the latent ability of a respondent and the characteristics of an item. We adapt this formulation to historical LLM-generated text detection by treating detectors as respondents, texts as items, and detector correctness as binary responses.

For historical text $X_i$ and detector $k$, define
$$R_{ik}=\mathbb{I}\{\hat{D}_{ik}=Y_i\},$$
where $Y_i$ is the true label and $\hat{D}_{ik}$ is the binary prediction produced by detector $k$. Thus, $R_{ik}=1$ indicates that detector $k$ correctly classifies text $i$, whereas $R_{ik}=0$ indicates an incorrect classification.

We use a multidimensional two-parameter logistic IRT model,
\begin{eqnarray}\label{irt}
P(R_{ik}=1|\theta_k,\alpha_i,\beta_i)=\frac{1}{1+\exp\{-\alpha_i^T(\theta_k-\beta_i)\}},
\end{eqnarray}
where $\alpha_i, \beta_i$, $\theta_k\in \mathbb{R}^d$, and $d$ is a prespecified latent dimension. Here $\theta_k$ denotes the latent ability vector of detector $k$, $\alpha_i$ is the discrimination vector of text $i$, and $\beta_i$ is the difficulty-related location  vector of text $i$. 

The discrimination vector $\alpha_i$ determines which latent dimensions are relevant to a text and how strongly its correctness probability changes with detector ability, whereas $\beta_i$ determines the item's location in the latent space. Because both quantities are multidimensional, difficulty should be interpreted through their joint projection rather than through individual elements of $\beta_i$. In particular, for fixed $\alpha_i$ and $\theta_k$, a larger value of $\alpha_i^T\beta_i$ lowers the probability of a correct response and therefore corresponds to greater effective difficulty. In this parameterization, the difficulty-related vector enters the response probability through the projection $\alpha_i^\top\beta_i$. Accordingly, the individual coordinates of $\beta_i$ are not separately identifiable from the response likelihood, and the fitted item representations are interpreted as regularized latent summaries rather than uniquely identified text characteristics.

The model parameters are estimated following \citet{maia_polo2024}, with implementation details provided in Appendix \ref{est_irt}. 
After fitting the model to the historical correctness matrix, the item parameters $Z_i=(\alpha_i^T,\beta_i^T)^T$ provide compact, response-based summaries of how each text interacts with the collection of detectors. These parameters characterize sample-level difficulty and discrimination in historical detector responses and constitute the IRT information used to construct the class-conditional detector-output likelihood in FIBER.

\section{Method}\label{sec3}

\subsection{Notation and data structure}

We first introduce the notation used throughout this section. Let $X_i$ denote the $i$-th historical text and $Y_i\in\{0,1\}$ its true label, where
$Y_i=1$ indicates machine-generated text and $Y_i=0$ indicates human-written text. 
Suppose that $K$ auxiliary detectors, excluding Fast-DetectGPT, are applied to each sample. Let
$\hat{D}_{ik}\in\{0,1\}$
denote the binary prediction of detector $k$, where $\hat{D}_{ik}=1$ indicates a machine-generated prediction and $\hat{D}_{ik}=0$ indicates a human-written prediction. The corresponding multi-detector output profile is $\hat{D}_i=(\hat{D}_{i1},\ldots,\hat{D}_{iK})^T$. 

For each labeled historical sample, detector correctness is defined as
$$R_{ik}=\mathbb{I}\{\hat{D}_{ik}=Y_i\}.$$
The historical correctness matrix $R=(R_{ik})$ is used to estimate the MIRT model introduced in Section \ref{sub_irt}.

Let $\theta_k\in\mathbb{R}^d$ denote the fitted ability vector of detector $k$, and let $Z_i=(\alpha^T_i, \beta^T_i)^T$ denote the resulting IRT item-parameter vector of text $X_i$, where $\alpha_i$ and $\beta_i$ are its discrimination and difficulty-related location vectors, respectively. In addition, let $s_i=s(X_i)$ denote the continuous Fast-DetectGPT score defined in Section \ref{sub_fast}. Because the continuous Fast-DetectGPT score is treated as a separate source of evidence, its thresholded binary decision is excluded from both $\hat{D}_i$ and $R$. This avoids using the same Fast-DetectGPT information twice in the proposed framework.

For standard FIBER, the historical information available after preprocessing is represented as 
$$\mathcal{D}_{sup}=\{(Y_i, s_i, \hat{D_i}, Z_i)\}_{i=1}^{n_{sup}}$$
together with the fitted detector ability parameters $\{\theta\}_{k=1}^K$.

For a new text $X_j$, its Fast-DetectGPT score $s_j$ and auxiliary detector-output profile $\hat{D}_j$ are observed, whereas its true label $Y_j$ is unknown. The corresponding IRT representation $Z_j$ is also unavailable because constructing detector correctness responses requires the true label. The practical inference objective is therefore
$$P(Y_j=1|s_j, \hat{D}_j; \mathcal{D}_{sup}).$$

FIBER integrates over the class-conditional distribution of the latent IRT representation learned from historical data. This allows historical detector–text interactions to contribute to inference through the observed detector-output pattern, without requiring test-specific IRT estimation.

\subsection{FIBER: Fast-DetectGPT with IRT-informed
Bayesian Evidence Recalibration}\label{fiber}

FIBER combines the continuous Fast-DetectGPT score with auxiliary detector outputs through Bayesian evidence recalibration. Its central component is an IRT-informed likelihood that captures historical patterns of detector correctness and their variation across texts. Since the IRT representation of an incoming text cannot be estimated without its true label, we treat it as a latent variable and integrate it out using class-conditional distributions learned from historical data. The resulting likelihood characterizes the observed detector-output pattern while accounting for dependence induced by shared latent text characteristics. This likelihood is then combined with the Fast-DetectGPT score and the historical class prior through Bayesian updating.

\subsubsection{IRT-informed detector-output likelihood} \label{detect}

Recall that the multidimensional two-parameter logistic IRT model defines the probability that detector \(k\) correctly classifies a text with latent representation $Z=(\alpha^T,\beta^T)^T$ as $$q_k(Z)=P(R_k=1|Z,\theta_k)=\frac{1}{1+\exp\{-\alpha^T(\theta_k-\beta)\}}.$$
Here,$q_k(Z)$ describes the correctness probability rather than the probability that detector \(k\) predicts the machine-generated class.

For $y\in \{0,1\}$, define
$$r_{yk}(Z)=P(\hat{D}_k=1|Z,Y=y)=(q_k(Z))^y(1-q_k(Z))^{(1-y)}.$$
For machine-generated texts, a positive detector prediction corresponds to a correct response. For human-written texts, the same positive prediction corresponds to an incorrect response.

Conditional on the class label and latent text representation, we assume that the auxiliary detector outputs are independent. The probability of observing a detector-output profile $\hat{D}$ is therefore
$$P(\hat{D}|Z,Y=y)=\prod_{k=1}^{K}
r_{yk}(Z)^{\hat{D}_k}[1-r_{yk}(Z)]^{1-\hat{D}_k}.$$

The latent representation \(\mathbf Z\) is not observed for incoming texts. To eliminate this unknown quantity, let $$f_y(Z)=f(Z|Y=y)$$ 
denote the class-conditional distributions of historical IRT item parameters. We define the IRT-informed detector-output likelihood as
$$h_y^{IRT}(\hat{D})=
\int
P(\hat{D}|Z,Y=y)f_y(Z)\textit{d}Z.$$
Unlike the conventional class-conditional Bernoulli model, the proposed likelihood allows detector correctness probabilities to vary with latent text characteristics. Although detector outputs are assumed conditionally independent given the latent representation and class label, marginalizing over the shared representation can induce dependence among their predictions. The resulting likelihood therefore incorporates historical information about detector ability and text-level heterogeneity, rather than relying solely on average detector performance.

Importantly, the IRT-informed likelihood is evaluated using the observed detector-output profile and historical model parameters. It does not require estimating IRT item parameters for incoming texts. This enables the historical IRT structure to contribute to sample-specific inference through the detector-output likelihood.

\subsubsection{Bayesian evidence recalibration}\label{bayes}

For a new text $X_j$, FIBER uses two sources of current evidence: its continuous Fast-DetectGPT score $s_j$ and its auxiliary detector-output profile $\hat{D}_j$. Define
$$g_y(s)=f(s|Y=y)$$ 
as the class-conditional Fast-DetectGPT score densities.

For the class prior, we use the empirical proportion of machine-generated texts in the labeled historical data, that is $\hat{\pi}=\frac{n_A}{n_{sup}}$, where $n_A$ is the number of historical machine-generated texts and $n_{sup}$ is the total number of labeled historical texts.

We assume that $s_j$ and $\hat{D}_j$ are conditionally independent given $Y_j$. Because these signals are derived from the same input text, residual dependence may remain even after conditioning on the class label. Consequently, the resulting posterior probabilities should be interpreted as model-based estimates rather than fully calibrated probabilities under arbitrary distribution shifts. Under this assumption, the posterior probability that text $X_j$ is machine-generated is 
\begin{eqnarray}\label{eq_real0}
\hat{p}_j=P(Y_j=1|s_j,\hat{D}_j; \mathcal{D}_{sup})=\frac{\hat{\pi}g_1(s_j)h_1^{IRT}(\hat{D}_j)}{\hat{\pi}g_1(s_j)h_1^{IRT}(\hat{D}_j)+(1-\hat{\pi})g_0(s_j)h_0^{IRT}(\hat{D}_j)}.
\end{eqnarray}
Equivalently, the posterior log-odds can be decomposed as
\begin{eqnarray}
logit(\hat{p}_j)= \log\frac{\hat{\pi}}{1-\hat{\pi}}+\log\frac{g_1(s_j)}
{g_0(s_j)}+\log\frac{h_1^{IRT}(\hat{D_j})}{h_0^{IRT}(\hat{D}_j)}.\label{sec_odd}
\end{eqnarray}
Although the IRT component accounts for dependence among auxiliary detector outputs induced by shared latent text characteristics, residual dependence between the Fast-DetectGPT score and the auxiliary detector outputs may remain. The conditional-independence assumption is therefore adopted as a practical approximation for evidence integration.

The three terms represent the empirical historical class prior, the current Fast-DetectGPT score evidence, and the IRT-informed detector-output evidence, respectively. Unlike a shared prior, the IRT likelihood ratio varies with the observed detector-output pattern of each incoming text and can therefore affect the relative ranking of predictions. This allows historical detector–text interactions to contribute directly to the discrimination of new texts.

In the standard FIBER configuration, the three components enter with unit weight. The implementation also allows nonnegative exponents to adjust their relative contributions, while the unweighted formulation corresponds to the Bayesian model described above.

The final prediction is obtained by 
$$\hat{Y}_j=\mathbb{I}\left\{\hat{p}_j \geq c\right\},$$
where $c \in [0,1]$ is a prespecified decision threshold.

\subsection{Parameter estimation}

The components of the proposed framework are estimated from labeled historical observations. These include the IRT model parameters, class-conditional distributions of historical IRT representations, Fast-DetectGPT score distributions, and the marginal detector-output likelihood.

\subsubsection{Estimation of IRT components and Fast-DetectGPT score distributions}\label{irt_est}
We first fit the multidimensional 2PL IRT model to the historical detector correctness matrix, obtaining detector ability vectors and text-level item representations. The two sets of parameters are estimated jointly and share the same latent coordinate system. To characterize the distribution of historical text characteristics within each class, we assume class-conditional multivariate Gaussian models for the fitted IRT representations.
\begin{eqnarray}\label{eq:z_norm}
Z|Y=1 \sim N\left(\mu_1^Z,\Sigma_1^Z\right), \quad
Z|Y=0 \sim N\left(\mu_0^Z,\Sigma_0^Z\right).
\end{eqnarray}
The class-specific means and covariance matrices are estimated from the fitted IRT item parameters of machine-generated and human-written historical texts by 
\begin{eqnarray}\label{eq:muz}
\hat{\mu}_y^Z
=
\frac{1}{n_y}
\sum_{i:Y_i=y} Z_i,
\end{eqnarray}
where $y\in\{0,1\}$, and $n_1=\sum_{i=1}^{n_{sup}}Y_i,
\quad
n_0=\sum_{i=1}^{n_{sup}}(1-Y_i)$, and the corresponding covariance estimates are 
\begin{eqnarray}\label{eq:sigza}
\hat{\Sigma}_y^Z
=
\frac{1}{n_y-1}
\sum_{i:Y_i=y}
(Z_i-\hat{\mu}_y^Z)(Z_i-\hat{\mu}_y^Z)^T.\end{eqnarray}
These estimates define the class-conditional distributions $f_y(Z)$, which are subsequently used to construct the IRT-informed detector-output likelihood through marginalization.

For the Fast-DetectGPT score, we assume class-conditional Gaussian distributions,
\begin{eqnarray}\label{eq:s_norm}
s|Y=y \sim N\left(\mu_y^s,(\sigma_y^s)^2\right).
\end{eqnarray}

For standard FIBER, the class-specific means and variances are estimated from the labeled historical Fast-DetectGPT scores. The resulting Gaussian density estimates $\hat{g}_y$ are subsequently evaluated at each incoming score $s_j$.

\subsubsection{Monte Carlo estimation of the IRT-informed detector likelihood}\label{muti_est}
The marginal detector-output likelihood involves integration over the \(2d\)-dimensional latent item representation. Because this integral does not generally admit a closed-form expression under the multidimensional logistic IRT model, we approximate it using Monte Carlo integration.

For each class $y\in\{0,1\}$, we generate $M=2{,}000$ independent Monte Carlo samples from its fitted IRT representation distribution: 
$$Z_y^{(m)}\sim N\left(
\hat{\mu}_y^Z,
\hat{\Sigma}_y^Z\right),
\qquad m=1,\ldots,M.$$
For each sampled representation, the fitted IRT model provides
$$\hat{q}_k(Z_y^{(m)})=
\frac{1}{
1+\exp\{
-(\alpha_y^{(m)})^T
(\hat{\theta}_k-\beta_y^{(m)})\}
}.$$
The corresponding probability of a positive detector output is
$$\hat{r}_{yk}^{(m)}=(\hat{q}_k(Z_y^{(m)}))^y[1-\hat{q}_k(Z_y^{(m)})]^{1-y}.$$

The class-conditional detector-output likelihood is then estimated as
$$
\hat{h}_y^{IRT}(\hat{D}_j)=
\frac{1}{M}
\sum_{m=1}^{M}
\prod_{k=1}^{K}
(\hat{r}_{yk}^{(m)})^{\hat {D}_{jk}}
(1-\hat{r}_{yk}^{(m)})^{1-\hat D_{jk}}.$$

The Monte Carlo samples are generated separately for the two classes. Once these samples have been drawn, the same class-specific samples can be reused to evaluate the likelihoods of multiple incoming texts.

The estimated prior, score densities, and IRT-marginalized detector likelihoods are combined using the posterior expression in Section \ref{bayes}. All historical model parameters are fitted using labeled training data, while each incoming text contributes only its current Fast-DetectGPT score and auxiliary detector-output profile. The fitted IRT parameters and class-conditional distribution estimates are treated as plug-in quantities during prediction. Consequently, the resulting posterior probabilities do not explicitly account for uncertainty in these historical parameter estimates. The complete FIBER procedure is summarized in  Algorithm \ref{alg1}.

\subsection{FIBER-Stream: summary-data inference with unlabeled score adaptation}
Standard FIBER assumes that labeled historical Fast-DetectGPT scores are available, allowing the class-conditional score distributions to be estimated directly. We further consider a more restrictive setting in which the original historical texts and their Fast-DetectGPT scores are no longer accessible. Only historical labels, detector-output profiles, and fitted IRT item parameters are retained. We refer to the corresponding extension as FIBER-Stream.

The retained historical information can be represented as
$$\mathcal{S}_{sup}=\{(Y_i,\hat{D}_i,Z_i)\}_{i=1}^{n_{sup}},$$ 
together with the fitted detector ability vectors $\{\hat{\theta}_k\}_{k=1}^K.$

These retained quantities are sufficient to estimate the historical class prior and the IRT-informed detector-output likelihoods described in Section \ref{detect}. However, because historical Fast-DetectGPT scores are unavailable, the class-conditional score densities cannot be estimated directly. 

For an incoming unlabeled batch of $n_{un}$ texts, the current Fast-DetectGPT scores and detector-output profiles are observed:
$$\mathcal{D}_{un}=\{(s_j,\hat{D}_j)\}_{j=1}^{n_{un}}.$$
FIBER-Stream estimates the unknown class-conditional score distributions from these unlabeled observations using an expectation-maximization (EM) algorithm, while keeping the historical prior and IRT-informed detector-output likelihoods fixed.

Under the same Gaussian score model as standard FIBER, the joint distribution of the observed score and detector-output profile is 
$$p(s_j,\hat{D}_j)=\hat{\pi}g_1(s_j)h_1^{IRT}(\hat{D}_j)+(1-\hat{\pi})g_0(s_j)h_0^{IRT}(\hat{D}_j).$$
The parameters of $g_1$ and $g_0$ are estimated from the incoming unlabeled batch.
Let $$\vartheta=(\mu_A^s, (\sigma_A^s)^2,\mu_H^s,(\sigma_H^s)^2)$$ denote the unknown score-distribution parameters. At EM iteration $t$, the $E$-step computes the posterior responsibility
\begin{eqnarray}\label{eq:omega}
\omega_j^{(t)}&=&P(Y_j=1|s_j,\hat{D}_j)\nonumber\\
&=&\frac{\hat{\pi}g_1(s_j,\vartheta^{(t)})h_1^{IRT}(\hat{D}_j)}{\hat{\pi}g_1(s_j,\vartheta^{(t)})h_1^{IRT}(\hat{D}_j)+(1-\hat{\pi})g_0(s_j,\vartheta^{(t)})h_0^{IRT}(\hat{D}_j)}.
\end{eqnarray}
%The corresponding expected complete-data log-likelihood is
%\begin{eqnarray}
%Q(\vartheta,\vartheta^{(t)})&=&\sum_{j=1}^{n_{un}}\omega_j^{(t)}\log(\hat{\pi}_{IRT}h_A(\hat{D}_j)g_A(s_j,\vartheta))\nonumber\\
%&&+\sum_{j=1}^{n_{un}}(1-\omega_j^{(t)})\log((1-\hat{\pi}_{IRT})h_H(\hat{D}_j)g_H(s_j,\vartheta)).
%\end{eqnarray}
The M-step treats these responsibilities as soft class assignments and updates
\begin{eqnarray}
\label{eq:emmua}\mu_A^{s,(t+1)}&=&\frac{\sum_{j=1}^{n_{un}}\omega_j^{(t)}s_j}{\sum_{j=1}^{n_{un}}\omega_j^{(t)}},\\
\label{eq:emsga}(\sigma_A^{s,(t+1)})^2&=&\frac{\sum_{j=1}^{n_{un}}\omega_j^{(t)}(s_j-\mu_A^{s,(t+1)})^2}{\sum_{j=1}^{n_{un}}\omega_j^{(t)}},
\\
\label{eq:emmuh}\mu_H^{s,(t+1)}&=&\frac{\sum_{j=1}^{n_{un}}(1-\omega_j^{(t)})s_j}{\sum_{j=1}^{n_{un}}(1-\omega_j^{(t)})},\\
\label{eq:emsgh}(\sigma_H^{s,(t+1)})^2&=&\frac{\sum_{j=1}^{n_{un}}(1-\omega_j^{(t)})(s_j-\mu_H^{s,(t+1)})^2}{\sum_{j=1}^{n_{un}}(1-\omega_j^{(t)})}.
\end{eqnarray}
To identify the two mixture components, we initialize the score-distribution means such that $\mu_A^s>\mu_H^s,$ consistent with the interpretation that larger Fast-DetectGPT scores provide stronger evidence for machine generation.

The E-step and M-step are repeated until convergence. Let $\hat{g}_A$ and $\hat{g}_H$ denote the resulting class-conditional score densities. The final machine-generated probability for text $j$ is
\begin{eqnarray}\label{eq:pred}
\hat{p}_j=P(Y_j=1|s_j,\hat{D}_j)=\frac{\hat{\pi}\hat{g}_1(s_j)h_1^{IRT}(\hat{D}_j)}{\hat{\pi}\hat{g}_1(s_j)h_1^{IRT}(\hat{D}_j)+(1-\hat{\pi})\hat{g}_0(s_j)h_0^{IRT}(\hat{D}_j)},
\end{eqnarray}
and the corresponding prediction is
$$\hat{Y}_j=\mathbb{I}\{\hat{p}_j\geq c\}.$$

FIBER-Stream therefore preserves the IRT-informed evidence-recalibration structure of standard FIBER while replacing supervised historical score estimation with EM-based adaptation from the incoming unlabeled batch. The historical IRT model, class-conditional item distributions, and detector-output likelihoods are reused without accessing the original historical texts. Because the score-distribution parameters are estimated jointly from an incoming batch rather than updated recursively for individual observations, FIBER-Stream should be interpreted as a batchwise or sequential inference procedure rather than fully online recursive learning. The complete FIBER-Stream procedure is summarized in Algorithm \ref{alg2}.

\section{Experiments and Results}\label{sec4}

We conduct three experiments to evaluate the FIBER family under complementary detection settings. First, we use an integrated multi-domain benchmark with 5-fold cross-validation to assess overall detection performance under a heterogeneous setting. Second, we analyze each domain separately to examine how detector performance changes across different text sources. Third, we consider an out-of-distribution black-box setting in which historical summaries learned from the integrated benchmark are transferred to an unseen target dataset without using target-domain labels for model estimation. This design progressively increases the difficulty of the evaluation, allowing us to assess the proposed framework from in-distribution performance to domain heterogeneity and, finally, cross-setting transfer under limited target information and distribution shift.

\subsection{Multi-domain data integration and detection}
We use the multi-domain benchmark constructed in our previous study \citep{zhao2026}, which contains 32,000 texts from arXiv, Writing Prompts, XSum, and Yelp. The dataset contains 16,000 human-written texts and 16,000 machine-generated texts produced by ChatGPT-5, Claude Sonnet 4.5, Qwen-3-32B, and Llama-3.3-70B. The four domains cover scientific writing, creative storytelling, news reporting, and online reviews, providing a heterogeneous setting in which both writing style and generator characteristics vary substantially.

Fourteen existing LLM-generated text detectors are evaluated, including 11 zero-shot detectors and 3 supervised detectors. The continuous Fast-DetectGPT score is retained as a separate text-level evidence source. Its thresholded binary prediction is therefore excluded from both the auxiliary detector-output matrix and the IRT correctness matrix, leaving $K=13$ auxiliary detectors. Detector correctness on the labeled historical samples is used to estimate the multidimensional IRT model, and the resulting item-parameter distributions are used to construct the IRT-informed detector-output likelihood. Detailed information on benchmark construction, generation settings, and detector implementation is provided in our previous study \citep{zhao2026}

All methods are evaluated using 5-fold cross-validation, with all trainable components estimated exclusively from the training set. This includes the IRT model, class-conditional distributions, detector-output likelihoods, and Weighted Voting coefficients. FIBER estimates the Fast-DetectGPT score distributions from labeled training scores, whereas FIBER-Stream treats these historical scores as unavailable and estimates the corresponding distributions from the unlabeled held-out batch through EM. Test labels are used only for final evaluation. In addition to the individual detectors, we include Hard Voting and Weighted Voting as conventional ensemble baselines. We further evaluate FIBER-Fast and FIBER-Stream-Fast, which exclude DetectGPT, DNA-GPT, and NPR from the 13 auxiliary detectors and retain the remaining 10 to examine how much performance can be preserved after removing several relatively expensive perturbation-, resampling-, or repeated-scoring-based detectors \citep{chen2025}.

\begin{table*}
\caption{Performance comparison of FIBER, FIBER-Stream, and the baseline methods under 5-fold cross-validation. Hard Voting denotes a simple majority-voting ensemble of the 14 baseline detectors, in which each text is assigned to the class receiving the larger number of detector votes. Values in parentheses indicate the rank of each method for the corresponding metric, with smaller ranks indicating better performance. Boldface highlights the proposed FIBER family.}\label{tab:met}
\centering
\begin{tabular}[!htbp]{@{}lcccc@{}}
\toprule
Detectors & Accuracy & Precision & Recall & F1 score\\
\midrule
\textbf{FIBER} & \textbf{0.620 (1)} & \textbf{0.642 (9)} & \textbf{0.542 (6)} & \textbf{0.587 (4)}\\
\textbf{FIBER-Fast} & \textbf{0.619 (2)} & \textbf{0.647 (8)} & \textbf{0.531 (7)} & \textbf{0.582 (6)}\\
\textbf{FIBER-Stream} & \textbf{0.606 (5)} & \textbf{0.616 (14)} & \textbf{0.565 (4)} & \textbf{0.589 (3)}\\
\textbf{FIBER-Stream-Fast} & \textbf{0.606 (5)} & \textbf{0.618 (13)} & \textbf{0.558 (5)} & \textbf{0.586 (5)}\\
Binoculars & 0.583 (13) & 0.676 (4) & 0.321 (15) & 0.435 (15)\\
DetectGPT & 0.618 (3) & 0.614 (15) & 0.632 (2) & 0.623 (1)\\
DNA-GPT & 0.563 (16) & 0.558 (19) & 0.612 (3) & 0.582 (6)\\
Entropy & 0.583 (13) & 0.630 (10) & 0.403 (13) & 0.491 (13)\\
Fast-DetectGPT & 0.587 (12) & 0.706 (2) & 0.301 (16) & 0.422 (16)\\
Likelihood & 0.615 (4) & 0.676 (4) & 0.444 (9) & 0.535 (8)\\
LogRank & 0.601 (8) & 0.662 (6) & 0.416 (11) & 0.510 (9)\\
LRR & 0.545 (17) & 0.630 (10) & 0.221 (18) & 0.327 (18)\\
NPR & 0.573 (15) & 0.600 (16) & 0.445 (8) & 0.509 (11)\\
Rank & 0.588 (11) & 0.630 (10) & 0.429 (10) & 0.510 (9)\\
Lastde & 0.594 (10) & 0.651 (7) & 0.406 (12) & 0.500 (12)\\
RoBERTa-ChatGPT & 0.520 (19) & 0.582 (17) & 0.164 (19) & 0.251 (19)\\
RoBERTa-Base & 0.524 (18) & 0.518 (20) & 0.692 (1) & 0.592 (2)\\
RoBERTa-Large & 0.505 (20) & 0.570 (18) & 0.043 (20) & 0.079 (20)\\
Hard Voting & 0.604 (7) & 0.695 (3) & 0.371 (14) & 0.484 (14)\\
Weighted Voting & 0.597 (9) & 0.797 (1) & 0.262 (17) & 0.394 (17)\\
\bottomrule
\end{tabular}
\end{table*}

Table \ref{tab:met} shows that the proposed methods perform particularly well in terms of overall accuracy. FIBER achieves the highest accuracy of 0.620, closely followed by FIBER-Fast at 0.619 and DetectGPT at 0.618. FIBER-Stream and FIBER-Stream-Fast both achieve an accuracy of 0.606. DetectGPT obtains the highest F1 score of 0.623, while RoBERTa-Base achieves the highest recall of 0.692 but with substantially lower precision. Among the proposed variants, FIBER-Stream achieves the highest F1 score of 0.589, followed by FIBER at 0.587. Compared with the original Fast-DetectGPT, FIBER increases accuracy from 0.587 to 0.620 and F1 score from 0.422 to 0.587, while recall improves substantially from 0.301 to 0.542 and precision decreases from 0.706 to 0.642. The gain therefore mainly comes from recovering a substantially larger proportion of machine-generated texts while retaining competitive precision.

The voting baselines help clarify why combining detectors is not sufficient by itself. Weighted Voting achieves the highest precision of 0.797 but its recall is only 0.262, resulting in an F1 score of 0.394. Hard Voting improves recall to 0.371 and F1 score to 0.484, but both voting approaches remain below the FIBER family in overall accuracy and F1 performance. These results show that combining detector decisions through equal or globally learned weights does not by itself resolve the imbalance in detector behavior. In contrast, FIBER combines the continuous Fast-DetectGPT score, the current auxiliary detector profile, and historical detector-response information within a probabilistic framework.

The reduced and summary-data variants also retain competitive performance. FIBER-Fast achieves an accuracy of 0.619 and an F1 score of 0.582, closely matching standard FIBER. FIBER-Stream reaches an accuracy of 0.606 and an F1 score of 0.589 without access to historical Fast-DetectGPT scores, while FIBER-Stream-Fast achieves corresponding values of 0.606 and 0.586 using only 10 auxiliary detectors. These results suggest that much of the useful information can be preserved when either the auxiliary detector set is reduced or only historical summary information is retained, although the relative performance varies across evaluation metrics.

\begin{figure}
    \centering
    \includegraphics[width=0.8\linewidth]{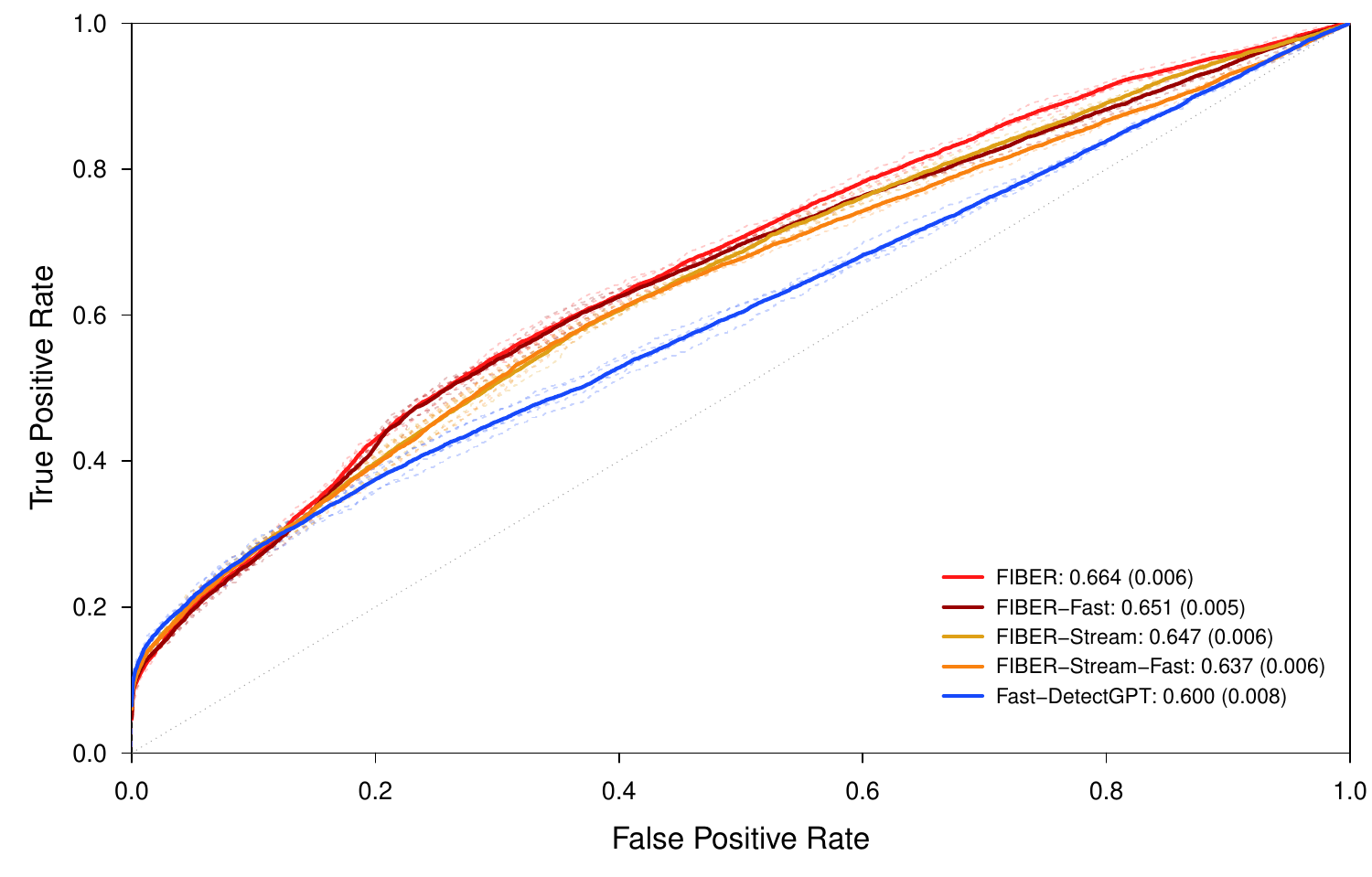}
    \caption{ROC curves for the four FIBER variants and Fast-DetectGPT under 5-fold cross-validation. The dashed curves represent fold-specific ROC curves, and the solid curves represent the average ROC curves across the five folds. The reported values are the mean AUCs, with the standard deviations of the fold-specific AUCs shown in parentheses.}
    \label{fig:roc}
\end{figure}

Figure \ref{fig:roc} provides a threshold-independent comparison of the four FIBER variants and Fast-DetectGPT. Standard FIBER increases the mean AUC from 0.600 for Fast-DetectGPT to 0.664, while FIBER-Fast achieves an AUC of 0.651. FIBER-Stream reaches 0.647 without access to historical Fast-DetectGPT scores, and FIBER-Stream-Fast achieves 0.637. The fold-specific AUC standard deviations are also small, ranging from 0.005 to 0.006 for the FIBER variants, compared with 0.008 for Fast-DetectGPT. All four FIBER configurations improve the threshold-independent ranking performance over Fast-DetectGPT, although the magnitude of improvement varies across the variants. Within the proposed framework, IRT provides a structured representation of historical detector–text interactions and enables the construction of a marginalized detector-output likelihood that accounts for latent text-level heterogeneity. These improvements reflect the performance of the integrated framework rather than the isolated contribution of IRT.

\subsection{Domain-specific data integration and detection}
The integrated benchmark combines four substantially different text sources and may therefore obscure domain-specific detector behavior. To examine this heterogeneity more directly, we repeat the evaluation separately on arXiv, Writing Prompts, XSum, and Yelp. Each domain contains 8,000 texts, equally divided between human-written and machine-generated samples. The same 5-fold cross-validation procedure is used, with all historical components and Weighted Voting coefficients estimated from the corresponding training folds. For the streaming variants, the class-conditional Fast-DetectGPT score distributions are estimated from the unlabeled held-out batch through EM.

Table \ref{tab:domain_result} reveals pronounced domain sensitivity among the individual detectors. Rank provides the clearest example: it achieves the highest F1 score of 0.790 on arXiv, but only 0.192 on Writing Prompts, 0.413 on XSum, and 0.140 on Yelp. NPR shows a similar pattern, with an F1 score of 0.784 on arXiv but substantially lower performance elsewhere. DetectGPT, in contrast, performs particularly well on Writing Prompts, where it achieves the highest accuracy of 0.722 and the highest F1 score of 0.710. On Yelp, Entropy becomes substantially stronger, achieving the highest recall of 0.654 and the highest F1 score of 0.602. More extreme domain dependence is observed for RoBERTa-Large and DNA-GPT. RoBERTa-Large exhibits nearly degenerate predictions on arXiv, with an accuracy of 0.500, a recall of only 0.001, and an undefined precision and F1 score, while DNA-GPT similarly collapses on XSum, yielding a recall of 0 and undefined precision and F1 score. These cases illustrate that distributional changes can not only alter the relative ranking of detectors but may cause some methods to produce highly imbalanced or effectively single-class predictions in particular domains. Overall, the results demonstrate that detector reliability depends strongly on the text domain, making it difficult to identify a single method that performs consistently well across heterogeneous settings.

\begin{table*}[htbp]
\caption{Domain-specific performance comparison of FIBER, FIBER-Stream, and the baseline methods under 5-fold cross-validation. Hard Voting denotes a simple majority-voting ensemble of the 14 baseline detectors, in which each text is assigned to the class receiving the larger number of detector votes. The best and second-best results for each metric within each domain are highlighted in bold and underlined, respectively. Boldface in the method column denotes the proposed FIBER family.}
\centering
\label{tab:domain_result}
\resizebox{\textwidth}{!}{%
\begin{tabular}[t]{@{}lcccc|cccc@{}}
\toprule
Domain & \multicolumn{4}{c}{arXiv} & \multicolumn{4}{c}{Writing Prompt}\\
\midrule
Detector & Accuracy & Precision & Recall & F1 score & Accuracy & Precision & Recall & F1 score\\
\midrule
\textbf{FIBER} & 0.751 & 0.759 & 0.735 & 0.747 & 0.633 & 0.712 & 0.448 & 0.549 \\
\textbf{FIBER-Fast} & 0.745 & 0.754 & 0.727 & 0.740 & 0.623 & 0.705 & 0.424 & 0.529 \\
\textbf{FIBER-Stream} & 0.707 & 0.712 & 0.699 & 0.704 & 0.638 & 0.708 & 0.471 & \underline{0.565} \\
\textbf{FIBER-Stream-Fast} & 0.706 & 0.696 & 0.735 & 0.714 & 0.631 & 0.708 & 0.447 & 0.547 \\
Binoculars & 0.621 & 0.660 & 0.505 & 0.571 & 0.634 & 0.700 & 0.469 & 0.562 \\
DetectGPT & 0.661 & 0.656 & 0.678 & 0.666 & \textbf{0.722} & 0.742 & \underline{0.680} & \textbf{0.710} \\
DNA-GPT & 0.643 & 0.643 & 0.642 & 0.642 & 0.597 & 0.682 & 0.364 & 0.474 \\
Entropy & 0.733 & 0.733 & 0.733 & 0.733 & 0.514 & 0.527 & 0.245 & 0.334 \\
Fast-DetectGPT & 0.624 & 0.643 & 0.565 & 0.599 & 0.643 & 0.740 & 0.445 & 0.555 \\
Likelihood & 0.747 & 0.736 & 0.769 & 0.752 & 0.583 & 0.686 & 0.320 & 0.433 \\
LogRank & 0.739 & 0.736 & 0.747 & 0.741 & 0.571 & 0.751 & 0.214 & 0.333 \\
LRR & 0.683 & 0.692 & 0.658 & 0.675 & 0.536 & 0.723 & 0.120 & 0.204 \\
NPR & \underline{0.788} & 0.799 & 0.770 & \underline{0.784} & 0.506 & 0.595 & 0.052 & 0.094 \\
Rank & \textbf{0.795} & \underline{0.808} & \underline{0.775} & \textbf{0.790} & 0.513 & 0.584 & 0.117 & 0.192 \\
Lastde & 0.670 & 0.677 & 0.651 & 0.663 & 0.640 & 0.739 & 0.432 & 0.545 \\
RoBERTa-Base & 0.509 & 0.505 & \textbf{0.890} & 0.644 & 0.540 & 0.715 & 0.132 & 0.222 \\
RoBERTa-Large & 0.500 & -- & 0.001 & -- & 0.501 & 0.492 & \textbf{0.787} & 0.543 \\
RoBERTa-ChatGPT & 0.572 & 0.638 & 0.335 & 0.438 & 0.527 & 0.565 & 0.236 & 0.332 \\
Hard Voting & 0.762 & 0.780 & 0.731 & 0.754 & 0.579 & \underline{0.769} & 0.228 & 0.351 \\
Weighted Voting & 0.787 & \textbf{0.878} & 0.666 & 0.757 & \underline{0.663} & \textbf{0.822} & 0.416 & 0.551 \\
\midrule
Domain & \multicolumn{4}{c}{XSum} & \multicolumn{4}{c}{Yelp}\\
\midrule
Detector & Accuracy & Precision & Recall & F1 score & Accuracy & Precision & Recall & F1 score\\
\midrule
\textbf{FIBER} & \underline{0.594} & 0.586 & \underline{0.655} & \underline{0.615} & \underline{0.591} & 0.596 & 0.568 & \underline{0.580} \\
\textbf{FIBER-Fast} & \textbf{0.596} & 0.594 & 0.626 & 0.605 & 0.586 & 0.606 & 0.525 & 0.552 \\
\textbf{FIBER-Stream} & 0.570 & 0.646 & 0.321 & 0.425 & 0.585 & 0.741 & 0.263 & 0.387 \\
\textbf{FIBER-Stream-Fast} & 0.568 & 0.668 & 0.281 & 0.392 & 0.583 & 0.738 & 0.258 & 0.381 \\
Binoculars & 0.565 & 0.706 & 0.223 & 0.339 & 0.578 & 0.729 & 0.249 & 0.371 \\
DetectGPT & 0.565 & 0.584 & 0.451 & 0.509 & 0.543 & 0.539 & \underline{0.599} & 0.567 \\
DNA-GPT & 0.500 & -- & 0.000 & -- & 0.568 & 0.605 & 0.395 & 0.477 \\
Entropy & 0.589 & 0.601 & 0.531 & 0.564 & 0.571 & 0.563 & \textbf{0.654} & \textbf{0.602} \\
Fast-DetectGPT & 0.565 & \underline{0.740} & 0.210 & 0.325 & 0.586 & 0.778 & 0.239 & 0.366 \\
Likelihood & 0.591 & 0.627 & 0.457 & 0.526 & 0.573 & 0.660 & 0.303 & 0.415 \\
LogRank & 0.578 & 0.637 & 0.380 & 0.470 & 0.555 & 0.699 & 0.201 & 0.307 \\
LRR & 0.534 & 0.606 & 0.196 & 0.296 & 0.533 & 0.700 & 0.116 & 0.198 \\
NPR & 0.517 & 0.536 & 0.332 & 0.399 & 0.495 & 0.539 & 0.465 & 0.370 \\
Rank & 0.545 & 0.581 & 0.320 & 0.413 & 0.514 & 0.607 & 0.079 & 0.140 \\
Lastde & 0.562 & 0.685 & 0.231 & 0.345 & 0.589 & 0.768 & 0.255 & 0.382 \\
RoBERTa-Base & 0.588 & 0.576 & \textbf{0.662} & \textbf{0.616} & 0.530 & 0.707 & 0.105 & 0.183 \\
RoBERTa-Large & 0.503 & 0.588 & 0.059 & 0.101 & 0.527 & 0.600 & 0.161 & 0.254 \\
RoBERTa-ChatGPT & 0.544 & 0.636 & 0.209 & 0.313 & 0.564 & \textbf{0.791} & 0.173 & 0.284 \\
Hard Voting & 0.562 & 0.694 & 0.224 & 0.338 & 0.570 & 0.774 & 0.198 & 0.313 \\
Weighted Voting & 0.560 & \textbf{0.768} & 0.173 & 0.283 & \textbf{0.593} & \underline{0.786} & 0.257 & 0.387 \\
\bottomrule
\end{tabular}%
}
\end{table*}

Against this pronounced domain heterogeneity, the FIBER family demonstrates relatively consistent and competitive performance across the four domains. Although it does not always achieve the best result, its performance remains competitive rather than exhibiting the severe domain-specific deterioration observed for several individual detectors. On arXiv, where Rank and NPR perform particularly well, FIBER achieves an accuracy of 0.751 and an F1 score of 0.747, only 0.044 and 0.043 below the corresponding best results of 0.795 and 0.790, respectively. On Writing Prompts, although DetectGPT achieves the highest F1 score of 0.710, FIBER-Stream obtains the second-highest F1 score of 0.565, while standard FIBER reaches 0.549. In contrast, Rank and NPR, which achieve F1 scores of 0.790 and 0.784 on arXiv, decline sharply to 0.192 and 0.094 on Writing Prompts, respectively. On XSum, FIBER-Fast achieves the highest accuracy of 0.596, while standard FIBER obtains an F1 score of 0.615, nearly matching the best result of 0.616. A similar pattern is observed on Yelp, where FIBER achieves an accuracy of 0.591 and an F1 score of 0.580, close to the corresponding best results of 0.593 and 0.602. Notably, the F1 scores of standard FIBER range from 0.549 to 0.747 across the four domains, whereas those of Rank and NPR range from 0.140 to 0.790 and from 0.094 to 0.784, respectively. These results suggest that, while the proposed framework is not uniformly optimal, it provides a more balanced performance profile across heterogeneous domains and avoids the pronounced performance deterioration exhibited by some individual detectors.

Weighted Voting achieves the highest precision on arXiv, Writing Prompts, and XSum, with values of 0.878, 0.822, and 0.768, respectively. It also performs competitively in accuracy on arXiv and Writing Prompts. However, its recall declines to 0.416 on Writing Prompts and only 0.173 on XSum, leading to substantially lower F1 scores in these settings. On Yelp, Weighted Voting achieves an accuracy of 0.593 and precision of 0.786, but its recall remains only 0.257. These results illustrate the limitation of summarizing detector reliability through a single set of global weights when the relative behavior of the detectors varies considerably across samples and domains.

The streaming variants provide a more restrictive test of the proposed framework. FIBER-Stream achieves F1 scores of 0.704, 0.565, 0.425, and 0.387 on arXiv, Writing Prompts, XSum, and Yelp, respectively. FIBER-Stream-Fast obtains corresponding F1 scores of 0.714, 0.547, 0.392, and 0.381. Standard FIBER-Stream outperforms Fast-DetectGPT in F1 score across all four domains, although its accuracy is slightly lower on Writing Prompts and Yelp. FIBER-Stream-Fast also improves upon Fast-DetectGPT in F1 score on arXiv, XSum, and Yelp, but performs slightly worse on Writing Prompts. The reduced performance of the streaming variants, particularly on XSum and Yelp, is consistent with the greater difficulty of estimating class-conditional Fast-DetectGPT score distributions from unlabeled batches in these domains. Nevertheless, these variants retain useful detection performance without requiring access to historical Fast-DetectGPT scores.

\begin{figure}[t]
    \centering
    \includegraphics[width=1\linewidth]{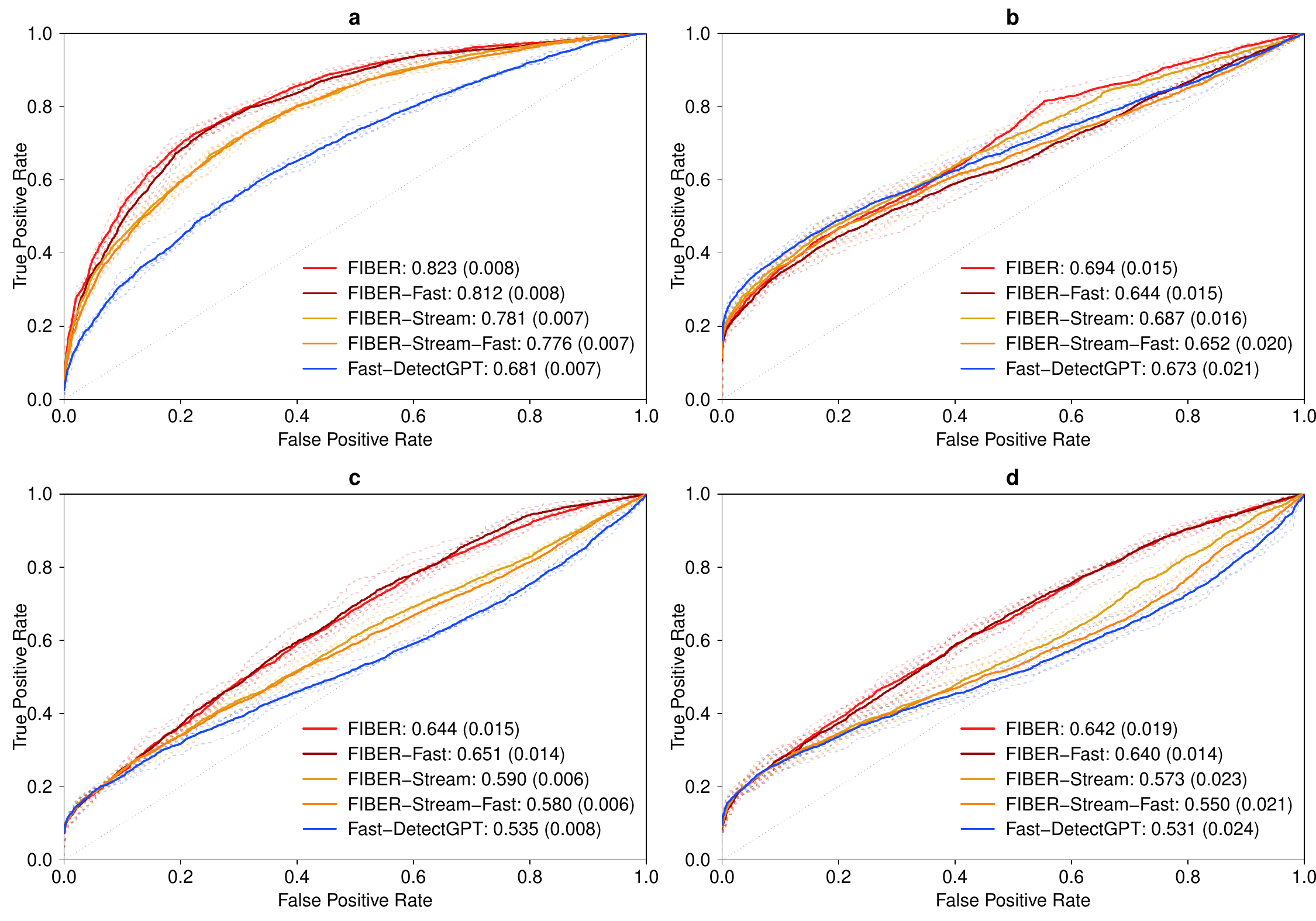}
    \caption{Domain-specific ROC curves for the four FIBER variants and Fast-DetectGPT under 5-fold cross-validation: (a) arXiv, (b) Writing Prompts, (c) XSum, and (d) Yelp. The dashed curves represent fold-specific ROC curves, and the solid curves represent the average ROC curves across the five folds. The reported values are the mean AUCs, with the standard deviations of the fold-specific AUCs shown in parentheses. }
    \label{fig:domain-roc}
\end{figure}

Figure \ref{fig:domain-roc} provides a threshold-independent comparison of the four FIBER variants and Fast-DetectGPT across individual domains. On arXiv, FIBER achieves the highest AUC among the plotted methods at 0.823, followed by FIBER-Fast at 0.812 and FIBER-Stream at 0.781, compared with 0.681 for Fast-DetectGPT. On Writing Prompts, FIBER again achieves the highest AUC at 0.694, followed by FIBER-Stream at 0.687 and Fast-DetectGPT at 0.673, whereas the reduced-detector variants obtain lower AUCs of 0.644 and 0.652. On XSum, FIBER-Fast and FIBER perform similarly, with AUCs of 0.651 and 0.644, respectively, compared with 0.535 for Fast-DetectGPT. A similar pattern appears on Yelp, where FIBER and FIBER-Fast achieve AUCs of 0.642 and 0.640, respectively, whereas Fast-DetectGPT reaches 0.531. Although the streaming variants show greater performance reductions on XSum and Yelp, they still achieve higher AUCs than Fast-DetectGPT in these two domains. Overall, the ROC results reinforce the domain-specific findings in Table 2, showing that the relative performance of different FIBER configurations varies across domains. Importantly, standard FIBER consistently improves AUC over Fast-DetectGPT across all four domains, demonstrating the benefit of integrating complementary detector evidence under heterogeneous text distributions.

\subsection{Out-of-distribution evaluation under black-box generation settings}
The first two experiments estimate the historical components from training folds drawn from the same benchmark as the evaluation data. We next consider a more demanding transfer setting in which the target generator and data distribution differ from those used to construct the historical benchmark. We use an external benchmark derived from the Fast-DetectGPT study \citep{bao2023fast}, covering XSum, Writing Prompts, and PubMedQA. Machine-generated texts are produced by ChatGPT/gpt-3.5-turbo and GPT-4 with temperature 0.8, providing a black-box setting involving source models and generation configurations not represented in the historical benchmark.

For each domain and source model, 150 machine-generated texts are included, yielding 900 machine-generated samples. Because the same human-written text may be paired with outputs from multiple source models, duplicate human-written samples are removed within each domain. The final evaluation set contains 812 human-written and 900 machine-generated texts, resulting in a total of 1,712 samples.

\begin{table*}
\caption{Performance comparison of FIBER, FIBER-Stream, and baseline methods on the out-of-distribution black-box benchmark. Hard Voting denotes a simple majority-voting ensemble of the 14 baseline detectors. The best and second-best values for each metric are highlighted in bold and underlined, respectively; boldface in the method column denotes the FIBER family.}
\centering\label{tab:result_bao}
\begin{tabular}[!h]{@{}lcccc@{}}
\toprule
 Detector & Accuracy & Precision & Recall & F1 score\\
\midrule
\textbf{FIBER}&0.827 &\underline{0.888}&0.768&0.824\\
\textbf{FIBER-Fast}&0.811&0.818&0.823&0.821\\
\textbf{FIBER-Stream}&\underline{0.840}&0.885
&0.800&\underline{0.840}\\
\textbf{FIBER-Stream-Fast}&0.827&0.839&\underline{0.831}&0.835\\
Binoculars & \textbf{0.859}& \textbf{0.903} & 0.820  & \textbf{0.860} \\
DetectGPT & 0.633  & 0.641 & 0.688 & 0.663 \\
DNA-GPT & 0.701  & 0.673  & \textbf{0.839} & 0.747 \\
Entropy & 0.397  & 0.441 & 0.550 & 0.490 \\
Fast-DetectGPT & 0.835& 0.875& 0.800 & 0.836\\
Likelihood & 0.765& 0.803& 0.733 & 0.767\\
LogRank & 0.762 & 0.792& 0.741 & 0.766\\
LRR & 0.693 & 0.715 & 0.690& 0.702\\
NPR & 0.641& 0.678& 0.602& 0.638 \\
Rank & 0.637& 0.679 & 0.587& 0.629\\
Lastde & 0.799  & 0.854& 0.746& 0.796\\
RoBERTa-Base & 0.644  & 0.673 & 0.629 & 0.650 \\
RoBERTa-Large & 0.602& 0.645 & 0.540 & 0.588 \\
RoBERTa-ChatGPT & 0.626  & 0.779 & 0.403  & 0.531\\
Hard Voting&0.793&0.868&0.716 &0.784\\
Weighted Voting&0.738 &0.822 &0.642 &0.721\\
\bottomrule
\end{tabular}
\end{table*}

In this experiment, the historical components are estimated using the full multi-domain benchmark of 32,000 labeled texts, without cross-validation. Standard FIBER transfers the historical class prior, fitted IRT parameters, IRT-informed detector-output likelihoods, and class-conditional Fast-DetectGPT score distributions to the target benchmark. FIBER-Stream retains the historical class prior and IRT-informed detector-output likelihoods but estimates the class-conditional Fast-DetectGPT score distributions from the unlabeled target batch through EM. The reduced variants follow the same respective procedures using 10 auxiliary detectors, while Weighted Voting transfers coefficients fitted on the source benchmark. No target-domain labels are used to estimate these model components or perform EM adaptation. For consistency with the baseline evaluation protocol, however, decision thresholds for methods producing continuous scores are selected retrospectively by maximizing Youden's index on the labeled OOD test set; methods evaluated using only binary predictions retain their original decisions. Consequently, the reported accuracy, precision, recall, and F1 scores represent post-hoc threshold-calibrated performance, whereas AUC is threshold-independent.

Table \ref{tab:result_bao} shows that the proposed FIBER family maintains competitive performance under distribution shift. Although none of the proposed variants achieves the best result for an individual metric, the FIBER family consistently attains the second-best performance across all four classification metrics, with relatively small differences from the corresponding best results. Specifically, Binoculars achieves the highest accuracy of 0.859, precision of 0.903, and F1 score of 0.860, while DNA-GPT obtains the highest recall of 0.839. Among the proposed methods, FIBER-Stream achieves the second-highest accuracy of 0.840 and F1 score of 0.840, only 0.019 and 0.020 below the corresponding best results. Standard FIBER obtains the second-highest precision of 0.888, compared with the best value of 0.903, whereas FIBER-Stream-Fast achieves the second-highest recall of 0.831, approaching the best result of 0.839. These findings indicate that, despite substantial changes in the target distribution, the FIBER family remains close to the strongest competing methods across different evaluation metrics rather than experiencing severe performance deterioration.

The streaming and computationally reduced variants further demonstrate the flexibility of the proposed framework. FIBER-Stream enables detection when historical Fast-DetectGPT scores are unavailable by estimating the class-conditional score distributions from unlabeled target data, while the reduced variants require fewer auxiliary detectors. These results suggest that FIBER can accommodate different historical data availability and computational requirements while preserving useful detection performance.

The voting baselines provide an additional comparison. Hard Voting achieves an accuracy of 0.793 and an F1 score of 0.784, whereas Weighted Voting obtains 0.738 and 0.721, respectively. Both methods perform below the FIBER family in these two metrics. These results illustrate that conventional aggregation through equal or globally estimated detector weights may be insufficient when detector reliability changes across settings. In contrast, FIBER combines historical detector-response information with the continuous Fast-DetectGPT score and the auxiliary detector-output profile of each incoming text.

\begin{figure}
    \centering
    \includegraphics[width=0.8\linewidth]{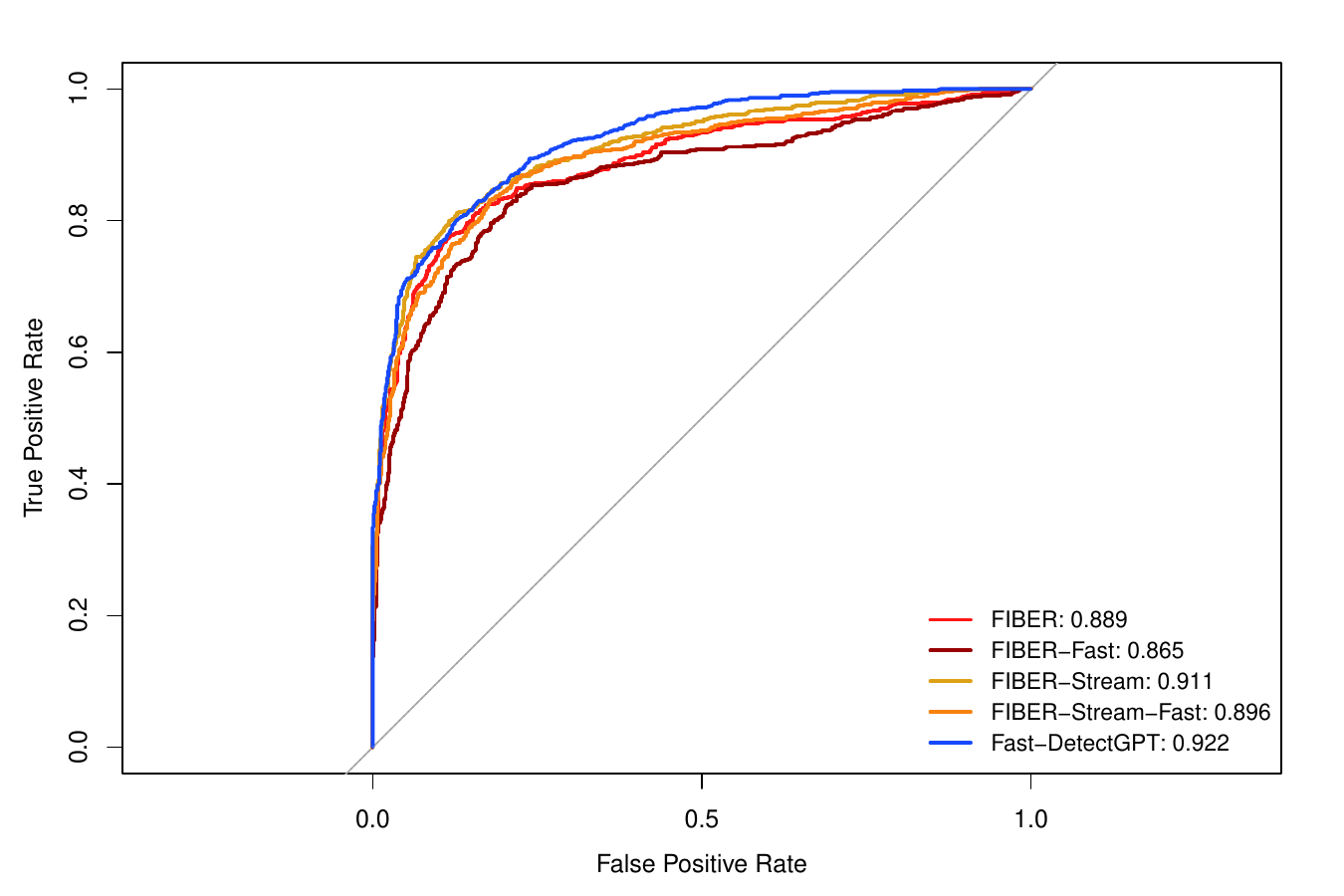}
    \caption{ROC curves for the four FIBER variants and Fast-DetectGPT on the out-of-distribution black-box benchmark. The reported values denote the corresponding AUCs.}
    \label{fig:roc_bao}
\end{figure}

Figure \ref{fig:roc_bao} provides a threshold-independent comparison under distribution shift. Among the five methods compared in Figure \ref{fig:roc_bao}, Fast-DetectGPT achieves the highest AUC of 0.922, while FIBER-Stream obtains the second-highest AUC of 0.911, with a difference of only 0.011. FIBER-Stream-Fast follows with an AUC of 0.896, compared with 0.889 for standard FIBER and 0.865 for FIBER-Fast. Notably, both streaming variants improve AUC over their corresponding standard counterparts, suggesting that unlabeled target-batch adaptation can improve ranking performance within the proposed framework. Although the FIBER variants do not surpass Fast-DetectGPT in AUC on this benchmark, the relatively small gap between FIBER-Stream and the best-performing method further supports the competitiveness of the proposed approach under distribution shift.

Taken together, the three experiments show that the strongest individual detector varies considerably across domains and evaluation settings. Rather than consistently achieving the best result for every metric, the FIBER family maintains a balanced and competitive performance profile, frequently approaching the strongest methods without exhibiting the pronounced deterioration observed for some individual detectors. In particular, its near-best performance across multiple metrics in the OOD experiment, together with the consistent AUC improvements in the domain-specific experiments, suggests that integrating current detection evidence with reusable historical detector-response information provides a useful strategy for detection across heterogeneous settings.

\section{Conclusion}\label{sec5}
In this paper, we propose FIBER, a Bayesian evidence-recalibration framework for LLM-generated text detection that combines the continuous Fast-DetectGPT score with auxiliary detector outputs and historical detector-response information. Rather than summarizing historical detector performance only through global accuracy or ensemble weights, FIBER applies item response theory to historical detector correctness patterns and uses the class-conditional distributions of the resulting item representations to construct an IRT-informed detector-output likelihood. By marginalizing over the unobserved IRT representations, FIBER incorporates historical detector–text interactions into sample-specific Bayesian inference without requiring test-specific IRT parameter estimation. We further develop FIBER-Stream for settings in which the original historical texts and their Fast-DetectGPT scores are unavailable, combining retained historical summaries with EM-based estimation from incoming unlabeled batches.

The experimental results demonstrate the effectiveness and competitiveness of the proposed framework across heterogeneous detection settings. On the integrated multi-domain benchmark, FIBER achieves the highest accuracy of 0.620 and improves the mean AUC from 0.600 for Fast-DetectGPT to 0.664. The domain-specific experiments further reveal substantial variation in the performance of individual detectors across arXiv, Writing Prompts, XSum, and Yelp. In contrast, FIBER maintains a relatively balanced performance profile and consistently improves AUC over Fast-DetectGPT across all four domains, without exhibiting the pronounced performance deterioration observed for some individual detectors. Under out-of-distribution transfer, the FIBER family achieves the second-best results across accuracy, precision, recall, and F1 score under retrospective threshold calibration, with only small differences from the corresponding best-performing methods. FIBER-Stream also achieves the second-highest AUC of 0.911 among the methods included in the ROC comparison, approaching the best result of 0.922 obtained by Fast-DetectGPT. Moreover, the streaming and computationally reduced variants retain competitive performance under more restrictive data-availability and detector configurations. Together, these findings suggest that the main advantage of FIBER lies not in consistently outperforming every individual detector, but in integrating complementary current evidence with reusable historical detector information to achieve competitive and relatively balanced performance across different detection settings.

Several directions remain for future research. The current framework relies on Gaussian and conditional-independence assumptions, and the specific contribution of IRT-based modeling relative to simpler detector-output likelihoods remains to be further investigated. In addition, FIBER-Stream currently employs batchwise adaptation, while the out-of-distribution classification results are based on retrospective threshold calibration. Future work could explore more flexible evidence models and fully sequential updating strategies to further improve adaptability and robustness under distribution shift.
\backmatter
\section*{Statements and Declarations}
\textbf{Competing Interests.} The authors declare that they have no competing interests.

\section*{Data Availability} The selected benchmark data and code are available at https://github.com/YbinZhao/FIBER.

\section*{Acknowledgements}
This work was partially supported by Shanghai Polytechnic University under Grant No. EGD26QD18 and the National Natural Science Foundation of China (NSFC) under Grant No. 12201158.

\vskip 0.2in
\bibliography{Fiber}

\newpage

\begin{appendices}
\counterwithin{figure}{section}
\counterwithin{table}{section}

\renewcommand{\thefigure}{\thesection\arabic{figure}}
\renewcommand{\thetable}{\thesection\arabic{table}}

\section{Estimation of IRT Parameters}\label{est_irt}

We estimate the IRT parameters from the historical detector correctness matrix defined in Section \ref{sub_irt}. Let $i=1,\cdots,I$ index texts, which are treated as items, and let $k=1,\cdots,K$ index detectors, which are treated as respondents. The binary response $R_{ik}=1$ indicates that detector $k$ correctly classifies text $i$, and $R_{ik}=0$ otherwise.

Following \citet{maia_polo2024}, we use a multidimensional two-parameter logistic IRT model.
$$p_{ik}\triangleq P(R_{ik}=1|\theta_k,\alpha_i,\beta_i) = \frac{1}{1+\exp\{-\alpha_i^T(\theta_k-\beta_i)\}},$$
where $\theta_k\in\mathbb{R}^{d}$ denotes the latent ability vector of detector $k$, $\alpha_i\in \mathbb{R}^d$ denotes the discrimination vector of text $i$, and $\beta_i\in\mathbb{R}^d$ denotes its difficulty-related location vector. The discrimination parameters characterize how strongly detector correctness changes along the latent ability dimensions, whereas the difficulty-related location parameters determine the position of the text in the latent space. As discussed in Section \ref{sub_irt}, difficulty in the multidimensional model is interpreted jointly through $\alpha_i^T\beta_i$, rather than through individual elements of $\beta_i$.

Following the hierarchical Bayesian specification used in the IRT implementation, Gaussian priors are assigned to the detector ability and text-specific item parameters, and model estimation is performed using variational inference \citep{zhao2026}. Posterior means are used as point estimates of $\theta_k$, $\alpha_i$, and $\beta_i$. We set the latent dimension to $d=5$ throughout all experiments.

During cross-validation, the IRT model is fitted independently using only the labeled training fold, with detector ability and item parameters estimated in a shared latent coordinate system. In the multidimensional setting, the difficulty-related vector $\beta_i$ represents a relative location, while the discrimination vector $\alpha_i$ characterizes how detector correctness varies across latent ability dimensions. Since our primary objective is predictive evidence integration rather than individual parameter interpretation, the fitted item parameters are concatenated as $Z_i=(\alpha_i^T,\beta_i^T)^T$. Their class-conditional distributions are modeled using multivariate Gaussian distributions with full covariance matrices to capture the empirical dependence between discrimination and difficulty-related parameters. These fitted representations are subsequently used to construct the IRT-informed detector-output likelihood through marginalization.

\section{Algorithms for FIBER and FIBER-Stream}
\label{app:Algorithm}
This appendix summarizes the computational procedures for FIBER and
FIBER-Stream. The two methods share the same Bayesian evidence-recalibration
structure but differ in the historical information available for estimating
the class-conditional Fast-DetectGPT score distributions. Standard FIBER uses
labeled historical scores to estimate these distributions, whereas
FIBER-Stream retains historical labels, auxiliary detector outputs, fitted
IRT item representations, and detector ability parameters, and estimates
the score distributions from an incoming unlabeled batch.

In both methods, the continuous Fast-DetectGPT score is treated as a
separate source of evidence. Its thresholded binary prediction is excluded
from the auxiliary detector-output profile and the historical correctness
matrix to avoid using the same score twice. Consequently, the standard
configuration uses $K=13$ auxiliary detectors out of the 14 evaluated
baseline detectors. The class prior is the empirical historical class
proportion, and the IRT item distributions contribute through the
marginalized detector-output likelihood, not through an IRT-informed prior.
\subsection{Algorithm 1 FIBER}\label{alg1}

\begin{breakablealgorithm}
\caption{FIBER}
\label{alg:fiber-observed-x}
\begin{algorithmic}[1]

\Require 
Labeled historical data 
$\mathcal{D}_{sup}=\{(X_i,Y_i,\hat D_i)\}_{i=1}^{n_{sup}}$;
incoming texts $\{X_j\}_{j=1}^{n_{un}}$ and their auxiliary detector-output profiles $\{\hat{D}_j\}_{j=1}^{n_{un}}$; Fast-DetectGPT scoring models; latent dimension $d$; Monte Carlo sample size $M$; decision threshold $c$.
\Ensure 
Posterior probabilities $\{\hat{p}_j\}_{j=1}^{n_{un}}$ and predicted labels
$\{\hat Y_j\}_{j=1}^{n_{un}}$.

\State Compute historical scores $s_i=s(X_i)$ and incoming scores $s_j=s(X_j)$.
\State Construct
$R_{ik}=\mathbb{I}\{\hat{D}_{ik}=Y_i\}$
for $i=1,\cdots,n_{sup}$ and $k=1,\cdots, K$.

\State Fit the multidimensional IRT model to $R$ and obtain $\{\hat{\theta}_k\}_{k=1}^K$ and $\hat{Z}_i=(\hat{\alpha}_i^T,\hat{\beta}^T_i)^T.$
\State Estimate the historical class prior $\hat{\pi}= \frac{1}{n_{sup}}\sum_{i=1}^{sup}Y_i$.
\For{$y\in\{0,1\}$}
    \State Estimate $\hat{f}_y(Z)=N(Z;\hat{\mu}_y^{Z},\hat{\Sigma}_y^Z)$ from $\{\hat{Z}_i:Y_i=y\}$.
    \State Estimate $\hat{g}_y(s)=N(s;\hat{\mu}_y^{s},\hat{\sigma}_y^{2s})$ from $\{\hat{s}_i:Y_i=y\}$.
    \For{$m=1,\ldots,M$}
    \State Draw $ Z_y^{(m)}=(\alpha_y^{(m)},
      \beta_y^{(m)})\sim\hat f_y$.
    \For{$k=1,\cdots,K$}
      \State $q_{yk}^{(m)}=
      \{1+\exp[-(\alpha_y^{(m)})^{T}
      (\hat{\theta}_k-\beta_y^{(m)})]\}^{-1}$.
      \State $r_{yk}^{(m)}=(q_{yk}^{(m)})^y
      (1-q_{yk}^{(m)})^{1-y}$.
    \EndFor
  \EndFor
\EndFor
\For{$j=1,\ldots,n_{un}$}
  \For{$y\in\{0,1\}$}
    \State Compute the IRT-informed detector-output likelihood:
    \Statex \hspace{\algorithmicindent}$\displaystyle
    \hat h_y^{IRT}(\hat{D}_j)
      =\frac{1}{M}\sum_{m=1}^{M}\prod_{k=1}^{K}
      (r_{yk}^{(m)})^{\hat D_{jk}}
      (1-r_{yk}^{(m)})^{1-\hat D_{jk}}.$
  \EndFor
  \State Compute the posterior probability:
  \Statex \hspace{\algorithmicindent}$\displaystyle
    \hat p_j=
    \frac{\hat{\pi}\hat{g}_1(s_j)
      \hat{h}_1^{IRT}(\hat{D}_j)}
    {\hat{\pi}\hat{g}_1(s_j)
      \hat{h}_1^{IRT}(\hat{D}_j)
     +(1-\hat{\pi})\hat{g}_0(s_j)
      \hat{h}_0^{IRT}(\hat{D}_j)}.$
  \State Assign $\hat Y_j=\mathbb I\{\hat p_j\ge c\}$.
\EndFor
\end{algorithmic}
\end{breakablealgorithm}

\subsection{Algorithm 2 FIBER-Stream}\label{alg2}
\begin{breakablealgorithm}
\caption{FIBER-Stream}
\label{alg:fiber-unobserved-x}
\begin{algorithmic}[1]

\Require Retained historical summaries
 $\mathcal S_{sup}=\{(Y_i,\hat{D}_i,
 \hat{Z}_i)\}_{i=1}^{n_{sup}}$;
 fitted abilities $\{\hat{\theta}_k\}_{k=1}^{K}$;
 incoming unlabeled batch $\{X_j,\hat{D}_j\}_{j=1}^{n_{un}}$;
 Fast-DetectGPT scoring model; Monte Carlo sample size $M$;
 initial score parameters $vartheta^{(0)}$;
 EM tolerance and iteration limit; decision threshold $c$.
\Ensure Posterior probabilities $\{\hat p_j\}_{j=1}^{n_{un}}$
 and predicted labels $\{\hat Y_j\}_{j=1}^{n_{un}}$.
\State Estimate $\hat\pi= n_{sup}^{-1}
 \sum_{i=1}^{n_{sup}}Y_i$.
\For{$y\in\{0,1\}$}
  \State Estimate $\hat f_y(Z)=
   N(Z;\hat{\mu}_y^Z,
   \hat{\Sigma}_y^Z)$ from retained
   $\{\hat{Z}_i:Y_i=y\}$.
  \State Draw $Z_y^{(m)}\sim\hat f_y$,
   $m=1,\ldots,M$; evaluate $q_{yk}^{(m)}$ and $r_{yk}^{(m)}$
   for $k=1,\ldots,K$ as in Algorithm \ref{alg:fiber-observed-x}.
\EndFor
\State Compute the incoming scores $s_j=s(X_j)$,
 $j=1,\ldots,n_{\mathrm{un}}$.
\For{$j=1,\ldots,n_{\mathrm{un}}$}
  \For{$y\in\{0,1\}$}
    \State Compute and retain the fixed detector-output likelihood:
    \Statex \hspace{\algorithmicindent}$\displaystyle
      \hat h_{yj}=\hat h_y^{\mathrm{IRT}}(\hat{  D}_j)
      =\frac1M\sum_{m=1}^{M}\prod_{k=1}^{K}
      (r_{yk}^{(m)})^{\hat D_{jk}}
      (1-r_{yk}^{(m)})^{1-\hat D_{jk}}.$
  \EndFor
\EndFor
\State Initialize the two Gaussian score densities $g_y(s; \vartheta^{(0)})$;
 use an initialization with $\mu_{1}^{s,(0)}>\mu_{0}^{s,(0)}$.
\Repeat
  \State \textbf{E-step:} for each $j$, compute
  \Statex \hspace{\algorithmicindent}$\displaystyle
    \omega_j^{(t)}=
    \frac{\hat\pi\,g_1(s_j; \vartheta^{(t)})\,\hat h_{1j}}
    {\hat\pi\,g_1(s_j; \vartheta^{(t)})\,\hat h_{1j}
    +(1-\hat\pi)\,g_0(s_j; \vartheta^{(t)})\,\hat h_{0j}}.$
  \State \textbf{M-step:} update the class-specific means and variances
    in $ \vartheta^{(t+1)}$ by the responsibility-weighted
    formulas (9)--(12) in Section~3.4.
\Until{the convergence tolerance or maximum iteration limit is reached}
\State Denote the fitted score densities by $\hat g_1$ and $\hat g_0$.
\For{$j=1,\ldots,n_{\mathrm{un}}$}
  \State Compute the final posterior probability:
  \Statex \hspace{\algorithmicindent}$\displaystyle
    \hat p_j=
    \frac{\hat\pi\,\hat g_1(s_j)\,\hat h_{1j}}
    {\hat\pi\,\hat g_1(s_j)\,\hat h_{1j}
     +(1-\hat\pi)\,\hat g_0(s_j)\,\hat h_{0j}}.$
  \State Assign $\hat Y_j=\mathbb I\{\hat p_j\ge c\}$.
\EndFor
\end{algorithmic}
\end{breakablealgorithm}

\subsection{Computationally reduced variants}

FIBER-Fast and FIBER-Stream-Fast follow Algorithms \ref{alg:fiber-observed-x} and \ref{alg:fiber-unobserved-x} without changing the inference procedure. The only
difference is the auxiliary detector set: DetectGPT, DNA-GPT, and NPR are
excluded, leaving $K=10$ auxiliary detectors. The IRT model is fitted to
the correctness matrix of this reduced detector set, and the corresponding
detector abilities, class-conditional IRT distributions, detector-output
likelihoods, and posterior probabilities are all computed using these ten
detectors.

For cross-validation, decision thresholds are calibrated on the labeled
training folds and applied to the held-out texts. In the out-of-distribution
(OOD) experiment, the decision thresholds of continuous-score methods are
selected \emph{post hoc} by maximizing Youden's index using the labeled OOD
test set, consistently with the baseline evaluation protocol. Such target
labels are used for threshold selection and final metric calculation, but
not for fitting the source IRT model or historical distributions, nor for
FIBER-Stream's unlabeled-batch EM. Accordingly, threshold-based OOD metrics
represent retrospective test-set-optimized performance, whereas AUC does
not depend on the selected threshold.

\section{Hard Voting and Weighted Voting Baselines}
\label{app:voting}

To provide conventional multi-detector ensemble baselines, we consider Hard Voting and Weighted Voting \citep{ozdacs2026}. Both methods combine the binary predictions of the same 14 baseline detectors evaluated in the experiments, but differ in how the detector contributions are determined. Let $D_{ik}\in\{0,1\}$ denote the prediction of detector $k$ for text $i$, where $D_{ik}=1$ indicates a machine-generated prediction and $D_{ik}=0$ indicates a human-written prediction.

\subsection{Hard Voting}

Hard Voting assigns equal importance to all $K=14$ detectors. The ensemble prediction is defined as
$$\hat{Y}_{i}^{\mathrm{HV}}
=
\mathbb{I}
\left\{
\frac{1}{K}
\sum_{k=1}^{K}D_{ik}
>
\frac{1}{2}
\right\}.$$

Thus, a text is classified as machine-generated only when more than half of the detectors vote for the machine-generated class. Hard Voting requires no parameter estimation and assigns the same contribution to every detector regardless of differences in detector reliability.

\subsection{Weighted Voting}

Weighted Voting allows the contributions of individual detectors to be learned from labeled training data. To represent support for the two classes symmetrically, the binary detector outputs are first transformed into signed votes, $$S_{ik}=2D_{ik}-1,$$
where $S_{ik}=1$ indicates support for the machine-generated class and $S_{ik}=-1$ indicates support for the human-written class. The probability that text $i$ is machine-generated is modeled using logistic stacking,
$$p_i
=
P(Y_i=1\mid D_i)
=
\operatorname{logit}^{-1}
\left(
b+\sum_{k=1}^{K}w_kS_{ik}
\right),$$
where $b$ is an unconstrained intercept and $w_k\geq 0$ is the weight assigned to detector $k$. The nonnegativity constraint preserves the direction of each detector's original prediction while allowing detectors to contribute with different strengths.

The intercept and detector weights are estimated by minimizing the binary log loss on the labeled training data,
$$(\hat b,\hat{  w})
=
\arg\min_{b,  w}
-
\frac{1}{n_{\mathrm{tr}}}
\sum_{i=1}^{n_{\mathrm{tr}}}
\left[
Y_i\log(p_i)
+
(1-Y_i)\log(1-p_i)
\right],$$
subject to
$$w_k\geq 0,
\qquad k=1,\ldots,K.
$$

For a new text, the fitted probability is
$$\hat{p}_i
=
\operatorname{logit}^{-1}
\left(
\hat{b}+\sum_{k=1}^{K}\hat{w}_kS_{ik}
\right).$$
The resulting prediction for a new text is
$$\hat{Y}_{i}^{\mathrm{WV}}
=
\mathbb{I}
\left\{
\hat p_i\geq \tau
\right\},$$
where $\tau$ is the prespecified decision threshold.

For the integrated and domain-specific experiments, the Weighted Voting coefficients are estimated independently within each training fold and then applied unchanged to the corresponding held-out fold. Held-out labels are not used in weight estimation. In the out-of-distribution experiment, the coefficients estimated from the source multi-domain benchmark are transferred directly to the target benchmark without using target-domain labels.

Hard Voting and Weighted Voting use only the final binary decisions of the 14 baseline detectors. They therefore provide conventional ensemble references for evaluating whether detector aggregation alone is sufficient. In contrast, FIBER separates the continuous Fast-DetectGPT score from the auxiliary detector outputs and additionally incorporates class-conditional detector-output likelihoods and IRT-informed historical information.

\section{Experimental Protocol and Implementation Details}

For the integrated and domain-specific experiments, we use 5-fold cross-validation. Within each fold, all components requiring labeled data, including the IRT model, class-conditional distributions, historical class prior, IRT-informed detector-output likelihoods, detector-specific thresholds, and Weighted Voting coefficients, are estimated using only the training fold. For standard FIBER, the class-conditional Fast-DetectGPT score distributions are also estimated from the labeled training scores. The posterior classification threshold $c$ is selected by maximizing the Youden index on the training-fold ROC curve and is then applied unchanged to the corresponding held-out fold. Held-out labels are used only for final performance evaluation.

For FIBER-Stream, the held-out fold is treated as an unlabeled incoming batch. The historical class prior and IRT-informed detector-output likelihoods estimated from the training fold are retained, whereas the class-conditional Fast-DetectGPT score distributions are estimated from the unlabeled held-out scores through EM. No held-out labels are used during this estimation. The procedure should therefore be interpreted as batchwise transductive inference rather than strictly online sample-by-sample learning.

The EM algorithm is initialized with the constraint $\mu_A^{s,(0)}>\mu_H^{s,(0)}$, consistent with the orientation of the Fast-DetectGPT score. Iteration stops when the relative change in the observed-data log-likelihood is below $\varepsilon=10^{-6}$ or when the maximum number of iterations $T_{max}=500$ is reached. FIBER-Fast and FIBER-Stream-Fast follow the same procedures as their standard counterparts but exclude DetectGPT, DNA-GPT, and NPR from the auxiliary detector set, leaving $K=10$ detectors.

\section{Out-of-Distribution Benchmark and Transfer Protocol}
The out-of-distribution benchmark is derived from the dataset released with Fast-DetectGPT \citep{bao2023fast} and contains texts from XSum, Writing Prompts, and PubMedQA, representing news writing, creative storytelling, and biomedical question answering, respectively. Machine-generated texts were produced by GPT-3.5-turbo and GPT-4 with a generation temperature of 0.8. For each domain–generator combination, 150 machine-generated texts are included, yielding 900 machine-generated samples. Because the same human-written text may appear in files associated with different generators, human-written samples are deduplicated within each domain using exact text matching. The final benchmark contains 812 human-written and 900 machine-generated texts, for a total of 1,712 samples.

No target-domain labels are used to estimate the historical model components. Standard FIBER transfers the historical class prior, fitted IRT parameters, IRT-informed detector-output likelihoods, and class-conditional Fast-DetectGPT score distributions estimated from the full multi-domain benchmark of 32,000 labeled texts. FIBER-Stream retains the historical class prior and IRT-informed detector-output likelihoods but estimates the class-conditional Fast-DetectGPT score distributions from the unlabeled target batch through EM. FIBER-Fast and FIBER-Stream-Fast follow the same respective procedures using the reduced auxiliary detector set.

For consistency with the baseline evaluation protocol, decision thresholds for methods producing continuous scores are selected retrospectively by maximizing Youden's index on the labeled OOD test set, while methods evaluated using only binary predictions retain their original decisions. Thus, target-domain labels are used for post-hoc threshold selection and performance evaluation, but not for fitting the historical IRT model, class-conditional distributions, or EM-based score adaptation. Accordingly, the reported threshold-dependent classification metrics represent test-set-calibrated performance, whereas AUC provides a threshold-independent measure of discrimination.
\end{appendices}

\end{document}